\documentstyle[12pt,a4,named]{article}
\begin{document}
\bibliographystyle{named}
\title{With raised eyebrows or the eyebrows raised ?
A Neural Network Approach to Grammar Checking for Definiteness\thanks{
to appear in Proceedings of Nemlap-II, 15-18 September, Ankara, Turkey}}

\author{Gabriele Scheler\\
Institut f{\"u}r Informatik\\ TU M{\"u}nchen\\ D-80290 M{\"u}nchen\\
scheler@informatik.tu-muenchen.de}
\maketitle
\begin{abstract}
In this paper, we use a feature model of the semantics of
plural determiners to present an approach to grammar checking for definiteness.
Using neural network techniques, a 
semantics -- morphological category mapping was learned.
We then applied a textual encoding technique
to the 125 occurences of the relevant category in a 10 000 word narrative text 
and learned a surface -- semantics mapping.
By applying the learned generation function to the newly generated
representations, we achieved a correct category assignment in many cases (87\%). 
These results are considerably better than
a direct surface categorization approach (54 \%), with a baseline (always
guessing the dominant category) of 60 \%.
It is discussed, how these results could be used in multilingual
NLP applications.
\end{abstract}
\section{Introduction}
Most uses of the definiteness category in English are grammatically constrained,
i.e. a substitution of a definite for an indefinite determiner and vice versa
leads to ungrammatical sentences.
In this paper, we use a model of the semantics of
plural determiners to present an approach to automatic generation of 
the correct determiner. 
We have identified a set of semantic features for the description
of relevant meanings of plural definiteness. A small training
set (30 sentences) was created according to linguistic criteria, 
and a functional
mapping from the semantic feature representation to the overt
category of indefinite/definite article was learned using neural network
techniques.
We have then provided a surface-oriented textual encoding of a 10000 word
text corpus. 
We removed the target category in each 
relevant plural
noun occurrence, and automatically generated semantic representations from
the encoded text.
Because texts are semantically underdetermined, 
and the text encoding technique involves a further huge reduction of information
content,
these representations have some degree of noise. 
However, in generation we can
assign the correct category in many cases (87\%). 
These results are put into perspective with experiments 
on surface categorization of sentences, i.e. applying learning techniques
without the benefit of semantic representations.

The basic methodology in designing a semantic feature representation
consists in finding a set of semantic dimensions
which correspond to the logical distinctions expressed by a certain
grammatical category (cf.\ \cite{Kamp93,Link91a,Link91b,Scheler96}). In the case of definite determiners, we have
chosen the dimensions of givenness (i.e.\ type of anaphoric relation), 
of quantification, of type of reference (i.e.\ predication or denotation),
of boundedness (i.e.\ mass reference or individual reference),
and of collective agency. The different logical forms of the sentences can 
be represented by a set of sentential operators, which are defined
in first-order logic. These sentential operators can be used as atomic
semantic features, which are consequently sufficient in representing the
logical meaning of a sentence with respect to the chosen semantic dimensions.
This approach is significantly different from POS or sense-tagging systems
such as \cite{Yarowsky92,Schmid94,Brill93,Church88,Jelinek85}. 
A complete list of semantic features and dimensions is given in the appendix.
A semantic feature set is sufficient for the explanation of a given
morphological category if it is possible to generate this category from
the corresponding feature representation.

The paper is structured as follows:
First, we present an experiment in learning a generation function,
i.e. a mapping from semantic representations to surface categories.
Then we explain the principles of textual coding that we have used
for the semantic feature extraction experiments. Finally, we show how
these mapping functions can be combined to provide a grammar checker
for the definiteness category of English, and discuss possible applications
in multilingual NLP.

\section{From semantic features to morphological expression}
The question that has been investigated by the first experiment
is the adequacy of a semantic representation for noun phrases
which consists of the semantic dimensions and individual features 
given in the appendix. In particular, we wanted to know
how a functional assignment that has been learned by a set of 
linguistically chosen examples carries over to instances of the
relevant phenomenon in real texts.

\subsection{Method}
\label{method}
In order to answer this question, 
we use a connectionist method of supervised learning (``quickprop''
\cite{Fahlman88}, a variant
of the back-propagation algorithm), as implemented
in the SNNS-system (cf.\ \cite{SNNS}).
Supervised learning requires to set up a number of training examples,
i.e. cases, where both input and output of a function are given.
From these examples a mapping function is created, which generalizes
to new patterns of the same kind.

We created a small training corpus for typical occurrences of
bare plurals and definite plurals. Grammars written for second
language learning often provide a good possibility of obtaining a
small sample of individual sentences, designed to cover all
possible uses of a specific category in discourse.
30 example sentences with distinct feature representations were 
adapted from  \cite{Thompson/Martinet}. 
For these examples, semantic feature representation were
created by hand. Neutral values ($*$) were also included.
Inter-subject agreement of tagging of the data
was 94 \% for two subjects (myself and a student). I.e. there was 
disagreement for 37 tags (out of 625), most of which (22) concerned 
the category of anaphoric relation.

In principle, there is a better measure of judging the correctness
of the feature representation, as each of these features
refers to the logical interpretation of the sentence. 
This means that the feature representation can serve as an
intermediate step in creating a cognitive representation expressed
in first-order logic, in the same way as it has been realized
in \cite{Scheler/Schumann95} for aspectual categories.
Correctness may then be tested by creating a set of inferences
for each sentence. However, this work is only experimental at
present, and has not been performed for definite and indefinite noun
phrases yet.
Finally, the value of the chosen feature set and individual representations
becomes apparent, when we use these representations in the chosen
task of generating correct determiners for deliberately truncated
(i.e. minus the value for the target category) sentences.
 
\begin{table}
\begin{center}
\begin{tabular}{|l|}
\hline
	{\em He gives wonderful PARTIES.} \\
	{\tt new general predication pieces * } \\
	{\bf indef}
	\ \\
	{\em The MUSICIANS are practicing a new piece.}\\
	{\tt given all reference pieces collective } \\
	{\bf def}
	\ \\	
	{\em They were discussing BOOKS and the theater.}\\
	{\tt new general predicative mass * } \\
	{\bf indef}
	\ \\
\hline
\end{tabular}
\end{center}
\caption{Examples from the training set: Sentences, semantic representations,
and grammatical category}
\label{examples}
\end{table}

The symbolic descriptions were translated into binary patterns
using 1-of-n coding.
The assignment of the correct output category consisted in a binary
decision, namely, definite plural or bare (indefinite) plural. 

We wanted to know how such
a set of training examples relates to the patterns found
in real texts. Accordingly, we tested the acquired classification on a narrative text,
(``Cards on the table'' by A. Christie), for which the first 5 chapters
were taken, with a total of 
9332 words.
Every occurrence of a plural noun without a possessive or demonstrative
pronoun formed part of the dataset. Modification by a possessive pronoun
({\it my friends}), or a demonstrative pronoun ({\it those people}) leads
to a neutralization of the indefiniteness/definiteness distinction as
expressed by a determiner.
Generating possessive or demonstrative pronouns is beyond 
the goals of this research.
As a result, there
were 125 instances of definite or bare plural nouns. 
Of these, 75 instances had no determiner (the dominant category), and
50 instances had the determiner ``the''. This provides a baseline of
guessing  at 60\%.
For the text cases, another set of semantic representations was manually created.

\subsection{Results}
The mapping from semantics to grammatical category for the
example sentences could be learned perfectly, i.e. any semantic 
representation was assigned its correct surface category.

The learned classifier was then applied to the cases
derived from the running text.
A high percentage of correctness (97 \%) could be achieved
(cf. Table~\ref{learning}).

\begin{table}[htb]
\begin{center}
\setlength{\unitlength}{0.0075in}%
\begin{picture}(445,229)(20,555)
\thicklines
\put( 55,780){\line( 1, 0){ 10}}
\put(240,670){\framebox(0,0){}}
\put( 60,580){\line( 1, 0){405}}
\put(465,580){\line(-1, 0){  5}}
\put( 60,580){\line( 0, 1){200}}
\put( 60,580){\line( 0, 1){200}}
\put( 60,580){\line( 0, 1){200}}
\put( 80,585){\line( 0, 1){195}}
\put( 80,780){\line( 1, 0){ 40}}
\put(120,780){\line( 0,-1){200}}
\put( 55,770){\line( 1, 0){ 10}}
\put(160,585){\line( 0, 1){190}}
\put(160,775){\line( 1, 0){ 40}}
\put(200,775){\line( 0,-1){195}}
\put( 20,775){\makebox(0,0)[lb]{\raisebox{0pt}[0pt][0pt]{\footnotesize 100\%}}}
\put( 80,555){\makebox(0,0)[lb]{\raisebox{0pt}[0pt][0pt]{\footnotesize Learning}}}
\put(160,555){\makebox(0,0)[lb]{\raisebox{0pt}[0pt][0pt]{\footnotesize 
Generalization}}}
\put( 95,590){\makebox(0,0)[lb]{\raisebox{0pt}[0pt][0pt]{\footnotesize 30}}}
\put( 30,760){\makebox(0,0)[lb]{\raisebox{0pt}[0pt][0pt]{\footnotesize 97 \%}}}
\put(175,590){\makebox(0,0)[lb]{\raisebox{0pt}[0pt][0pt]{\footnotesize 121}}}
\end{picture}
\end{center}
\caption{Mapping from semantic representation to output category}
\label{learning}
\end{table}

This result is remarkable, as it involves a generalization from linguistically
selected, 'made-up' examples to real textual occurrences.
We may assume that the selected set of semantic features describes
the relevant semantic dimensions of the surface category
of definiteness.
We also examined
the few remaining misclassifications (cf. Table \ref{exceptions}). 
They are due
to stylistic peculiarities, as in {\em 45\/} and {\em 89}. 
Also, two sentences involving numerals
were not classified correctly. This has probably not been sufficiently 
covered by the training set.

\begin{table}
\begin{center}
\begin{tabular}{|l|}
\hline
{\em 45 INTRODUCTIONS completed, he gravitated naturally to the side of Colonel Race.} \\
{\tt given all predication mass collective }\\
{\bf indef}
\ \\
{\em 89 I held the most beautiful CARDS yesterday.} \\
{\tt new some predication pieces * }\\
{\bf def}
\ \\
{\em 94 He saw four EXPRESSIONS break up - waver.}\\ 
{\tt implied num predication pieces distributive} \\
{\bf indef}
\ \\
{\em 118 Yes. That's to say, I passed quite near him THREE TIMES.}\\
{\tt implied num predication pieces * }\\
{\bf indef}
\ \\
\hline
\end{tabular}
\end{center}
\caption{Misclassifications of the text cases}
\label{exceptions}
\end{table}

We have achieved to learn a generation function from semantic representations
with remarkably few wrong assignments. The remaining problems with 
functional assignment which are due to stylistic variation are
less than we expected, but they may go beyond an analysis in terms of 
semantic-logical features.

\section{Semantic feature extraction from Text}
For the goal of cognitive modeling it is interesting to look at
the kind of semantic representations necessary to explain attested 
morphological categories and their use.
For practical purposes, however, semantic representations cannot
be manually created. They have to be derived
from running text by automatic methods. This is a goal that is not 
easy to reach.

First of all, texts are semantically underdetermined. They do not
contain all the information present in a speaker's mind that
corresponds to a full logical representation. Fortunately, these
logical representations are often redundant for the selection
of a grammatical category, so that a noisy representation may be 
sufficient for practical NLP tasks such as text understanding, 
machine translation or grammar checking.
Secondly, there remains the problem of how to represent or {\it code}
a text such as to derive a maximum of semantic information from it,
but reduce its overall information content, which puts too much burden
on any current learning technique (in particular the large amount
of different lexical words).

In this paper we wanted to look at the possibility of using a neural
network learning approach to syntax-semantics mapping for {\it
grammar checking}, i.e. the automatic correction of the definiteness
category in a running text. This could be a valuable feature in a 
foreign language editor, it is also a significant part of any translation
system. 
\subsection{Text Encoding}
The text encoding technique should have two important properties:
\begin{itemize}
\item reducing the informational content of a text without losing its
essential parts for the task at hand 
\item using only readily accessible surface information, and limiting
pre-processing to a minimum
\end{itemize}
For the former goal we have provided representations using essentially
two syntactic schemas:\\
{\em NP -- predicate -- NP} and \\
{\em NP -- preposition -- NP}.

This is a fairly radical approach in reducing syntactic complexity,
and it is possible that more detailed representations of syntactic
relations would prove an asset in semantic feature extraction.
(alternative approaches to text encoding are contained in
\cite{BauerDIPLOM} and \cite{Scheler94g}). 
However the advantage of this simplistic scheme is that we can use a
single fixed-length slot-value representation which fits the local
context of most noun phrases.
The diversity of lexical items has been reduced by substituting each
lexical word by high-level syntactic-semantic features as derived from
WordNet \cite{Miller93}. Functional words and morphology have been
reduced to singular/plural and definite/indefinite distinctions.
The full textual encoding scheme looks as follows:
 
\begin{itemize}
\item[1.] head noun 
\begin{itemize}
\item[2.] adjectival/adverbial modifiers
\item[3.] number (singular/plural)
\item[4.] definiteness (indef/def or qu) 
\end{itemize}
\item[5.] predicate or preposition 
\item[6.] dependent noun 
\begin{itemize}
\item[7.] adjectival/adverbial modifiers
\item[8.] number (singular/plural)
\item[9.] definiteness (indef/def or qu) 
\end{itemize}
  
\end{itemize}
Values in the slots are lexical classes for
head noun, predicate and dependent noun (e.g.,
perceptual\_entity, physical\_object, body\_part, person, communication)
and grammatical classes for modifiers (e.g., adjective, numeral, demonstrative).
The difficult problem of word sense ambiguity which arises even on the level
of primary lexical classes, or syntactic-semantic features, was 
circumvented by assigning the most frequent lexical class to a 
lexical word, measured in terms of its different word senses.
An easy alternative, namely using all lexical classes in a distributed
lexical encoding, was not explored here.
Some examples are given in Table~\ref{input-examples}.
Using 1-of-n coding,  we get 53 bits (i.e. 53 features) in 9 slots.
We constructed another neural network with a 53-20-15 architecture
(input-hidden-output layer), where 20 hidden units proved to be optimal
for the given problem, and tried to learn a mapping function from the
surface encoding to the semantic layer (15 features) . 

\begin{table}
\begin{center}
\begin{tabular}{|l|}
\hline
{\em 3 VOICES drawled or murmured.} \\
{\tt perceptual\_entity * plural qu action * * * * } \\
\ \\
{\em 4 in aid of the London HOSPITALS. } \\
{\tt event * singular indef prep institution desc\_adj plural qu } \\
\ \\
{\em 5 a Lovely Young Thing with tight poodle CURLS.} \\
{\tt object desc\_adj singular indef prep body\_part desc\_adj plural qu} \\
\ \\
{\em 7 He wore a moustache with stiff waxed ENDS.} \\
{\tt body\_part * singular indef prep part desc\_adj plural qu} \\
\hline
\end{tabular}
\end{center}
\caption{Examples for surface textual coding}
\label{input-examples}
\end{table}

\subsection{Experiments}
In order to investigate the possibilities of grammar checking, we left out
the definiteness category for the target noun phrase, i.e. substituted
{\it indef/def} by {\it qu} for a single noun phrase per sentence. 

We have used crossvalidation by leaving-one-out for the 125
cases.
The number of examples is still fairly small for surface-semantics
mapping, accordingly we had to use the strong reduction in information
outlined above to have a noticeable generalization effect. In some
cases the resulting textual representations look alike, although there
are differences in semantic content, which is a major problem for
the learning technique used. A learning technique which would be less 
sensitive to conflicting data would probably improve the performance.
The results for learning and for generalization 
have been split up for the number of errors per pattern. They are
given in Table~\ref{input-learning}. 

\begin{table}[htb]
\begin{center}
\setlength{\unitlength}{0.0075in}%
\begin{picture}(445,265)(20,535)
\thicklines
\put( 55,780){\line( 1, 0){ 10}}
\put( 55,765){\line( 1, 0){ 10}}
\put( 55,650){\line( 1, 0){ 10}}
\put( 55,670){\line( 1, 0){ 10}}
\put( 80,580){\framebox(40,185){}}
\put(240,670){\framebox(0,0){}}
\put(240,580){\framebox(40,90){}}
\put(320,580){\framebox(40,70){}}
\put(400,580){\framebox(40,25){}}
\put( 60,800){\line( 0,-1){220}}
\put( 60,580){\line( 1, 0){405}}
\put(465,580){\line(-1, 0){  5}}
\put( 55,595){\line( 1, 0){ 10}}
\put(160,580){\framebox(40,15){}}
\put( 55,605){\line( 1, 0){ 10}}
\put( 20,775){\makebox(0,0)[lb]{\raisebox{0pt}[0pt][0pt]{\footnotesize 100\%}}}
\put( 20,757){\makebox(0,0)[lb]{\raisebox{0pt}[0pt][0pt]{\footnotesize 92\%}}}
\put( 20,665){\makebox(0,0)[lb]{\raisebox{0pt}[0pt][0pt]{\footnotesize 45\%}}}
\put( 20,647){\makebox(0,0)[lb]{\raisebox{0pt}[0pt][0pt]{\footnotesize 35\%}}}
\put(170,585){\makebox(0,0)[lb]{\raisebox{0pt}[0pt][0pt]{\footnotesize 10}}}
\put(250,585){\makebox(0,0)[lb]{\raisebox{0pt}[0pt][0pt]{\footnotesize 57}}}
\put(330,585){\makebox(0,0)[lb]{\raisebox{0pt}[0pt][0pt]{\footnotesize 43}}}
\put(410,585){\makebox(0,0)[lb]{\raisebox{0pt}[0pt][0pt]{\footnotesize 15}}}
\put( 90,590){\makebox(0,0)[lb]{\raisebox{0pt}[0pt][0pt]{\footnotesize 115}}}
\put( 80,555){\makebox(0,0)[lb]{\raisebox{0pt}[0pt][0pt]{\footnotesize Learning}}}
\put(160,555){\makebox(0,0)[lb]{\raisebox{0pt}[0pt][0pt]{\footnotesize correct}}}
\put(240,555){\makebox(0,0)[lb]{\raisebox{0pt}[0pt][0pt]{\footnotesize $\leq 2$ errors}}}
\put(320,555){\makebox(0,0)[lb]{\raisebox{0pt}[0pt][0pt]{\footnotesize $\leq 4$}}}
\put(400,555){\makebox(0,0)[lb]{\raisebox{0pt}[0pt][0pt]{\footnotesize $>4$ errors}}}
\put( 20,590){\makebox(0,0)[lb]{\raisebox{0pt}[0pt][0pt]{\footnotesize 8\%}}}
\put( 20,605){\makebox(0,0)[lb]{\raisebox{0pt}[0pt][0pt]{\footnotesize 12\%}}}
\put(250,535){\makebox(0,0)[lb]{\raisebox{0pt}[0pt][0pt]{\footnotesize Generalization}}}
\end{picture}
\end{center}
\caption{Mapping from encoded surface text to semantic representation}
\label{input-learning}
\end{table}

These results amount in a total average of 2.73 errors per pattern, where
15 bits had to be set. 
Our main goal was to generate a set of semantic feature representations
from sentences without target categories, and test how much
noise the previously learned generation function can tolerate.

\section{Grammar checking for determiners}

We have observed before that most uses of plural determiners in English
are grammatically constrained, and in many cases these grammatical
constraints are evident even from single sentences, without further textual 
context.
\subsection{Method}
In order to qualify whether a specific use of a determiner is sententially
constrained, we have given the list of 125 
sentences with the target categories changed to three native speakers.
We found that speakers agreed on a core of 15 sentences 
which were considered acceptable with the opposite category, and 
received a total of 22 sentences which at least one speaker
judged grammatical. This means, in 103 out of the 125 plural noun 
occurrences, speakers of English seem to have no choice in the use 
of the determiner.
By excluding the 22 sentences with 'free variation' we bypass problems 
of textual coreference and anaphora, which also play an important
role in determiner selection (cf.\ \cite{Aone-Bennett96}, \cite{Connolly95} for learning
approaches to anaphora resolution). Still the number of text cases that are
narrowly sententially constrained is fairly high.
For the remaining cases, we took the generalized semantic representations
from the previous experiment, and tested the performance with the
learned generation function.

\subsection{Results}

The results were encouraging: In many cases (89, i.e. 87 \%) the system
made the correct binary choice.
Note that these are generation data on representations that were derived
from unseen, only surface-encoded text. When we look at the relation
between error per pattern and generation performance 
(cf.~Table \ref{relation}),
a clear picture emerges. While the generation function is fault-tolerant
to a degree (app. $\leq$ 2 errors), its performance decreases when the
number of errors per pattern exceeds a certain limit ($>$ 2 errors), 
up to a point,
when we can only reproduce chance level ($>$4 errors). 

\begin{table}[htb]
\begin{center}
\setlength{\unitlength}{0.0075in}%
\begin{picture}(360,267)(20,555)
\thicklines
\put(120,785){\line( 0,-1){ 10}}
\put(115,780){\line( 1, 0){ 10}}
\put(240,745){\line( 0,-1){ 10}}
\put(235,740){\line( 1, 0){ 10}}
\put(300,670){\line( 0,-1){ 10}}
\put(295,665){\line( 1, 0){ 10}}
\put(180,775){\line( 0,-1){ 10}}
\put(175,770){\line( 1, 0){ 10}}
\put(210,760){\line( 0,-1){ 10}}
\put(205,755){\line( 1, 0){ 10}}
\put( 60,580){\vector( 0, 1){220}}
\put( 60,580){\vector( 1, 0){320}}
\put( 55,780){\line( 1, 0){ 10}}
\put(120,585){\line( 0,-1){ 10}}
\put(180,585){\line( 0,-1){ 10}}
\put(240,585){\line( 0,-1){ 10}}
\put(300,585){\line( 0,-1){ 10}}
\put(115,782){\line( 6,-1){ 60}}
\put(175,772){\line( 2,-1){ 60}}
\put(240,740){\line( 4,-5){ 60}}
\multiput( 55,680)(6.03960,0.00000){51}{\line( 1, 0){  3.020}}
\put(210,585){\line( 0,-1){ 10}}
\put(115,555){\makebox(0,0)[lb]{\raisebox{0pt}[0pt][0pt]{\footnotesize 0}}}
\put(360,555){\makebox(0,0)[lb]{\raisebox{0pt}[0pt][0pt]{\footnotesize no of errors per pattern}}}
\put( 20,775){\makebox(0,0)[lb]{\raisebox{0pt}[0pt][0pt]{\footnotesize 100\%}}}
\put( 20,810){\makebox(0,0)[lb]{\raisebox{0pt}[0pt][0pt]{\footnotesize correctness of generation}}}
\put(190,775){\makebox(0,0)[lb]{\raisebox{0pt}[0pt][0pt]{\footnotesize 96\%}}}
\put(210,725){\makebox(0,0)[lb]{\raisebox{0pt}[0pt][0pt]{\footnotesize 80\%}}}
\put(305,670){\makebox(0,0)[lb]{\raisebox{0pt}[0pt][0pt]{\footnotesize 42\%}}}
\put( 20,675){\makebox(0,0)[lb]{\raisebox{0pt}[0pt][0pt]{\footnotesize 50\%}}}
\put(235,555){\makebox(0,0)[lb]{\raisebox{0pt}[0pt][0pt]{\footnotesize $\leq 4$}}}
\put(295,555){\makebox(0,0)[lb]{\raisebox{0pt}[0pt][0pt]{\footnotesize $>4$}}}
\put(220,760){\makebox(0,0)[lb]{\raisebox{0pt}[0pt][0pt]{\footnotesize 87\% total}}}
\put(150,555){\makebox(0,0)[lb]{\raisebox{0pt}[0pt][0pt]{\footnotesize $\leq 2$}}}
\put(205,555){\makebox(0,0)[lb]{\raisebox{0pt}[0pt][0pt]{\footnotesize 2.7}}}
\end{picture}
\end{center}
\caption{Generation from automatically derived semantic representations}
\label{relation}
\end{table}

We may also compare this approach to a direct textual categorization
approach.
In this case we used the textual encoding (53 bits) and tried to learn
the morphological category by direct supervised learning. I.e.
instead of the 53-20-15 net for semantic feature extraction,
coupled by a 15-5-2 net for generation, we used a single
53-X-2 net (where X was optimal at 10) and repeated the learning
process for the 103 examples. The results were significantly
worse, they did not exceed chance level (cf. Table \ref{surface}).
Including extra hidden layers for automatic construction of a
``semantic layer'', i.e. a 53-20-10-15 net, did not significantly
improve these results (58/56\%)

\begin{table}[htb]
\begin{center}
\setlength{\unitlength}{0.0075in}%
\begin{picture}(445,229)(20,555)
\thicklines
\put( 55,780){\line( 1, 0){ 10}}
\put( 60,580){\line( 1, 0){405}}
\put(465,580){\line(-1, 0){  5}}
\put( 60,580){\line( 0, 1){200}}
\put( 60,580){\line( 0, 1){200}}
\put( 55,685){\line( 1, 0){ 10}}
\put( 60,580){\line( 0, 1){200}}
\put( 55,735){\line( 1, 0){ 10}}
\put( 80,585){\line( 0, 1){150}}
\put( 80,735){\line( 1, 0){ 40}}
\put(120,735){\line( 0,-1){150}}
\put(160,585){\line( 0, 1){100}}
\put(160,685){\line( 1, 0){ 40}}
\put(200,685){\line( 0,-1){105}}
\put( 20,775){\makebox(0,0)[lb]{\raisebox{0pt}[0pt][0pt]{\footnotesize 100\%}}}
\put( 80,555){\makebox(0,0)[lb]{\raisebox{0pt}[0pt][0pt]{\footnotesize Learning}}}
\put( 30,730){\makebox(0,0)[lb]{\raisebox{0pt}[0pt][0pt]{\footnotesize 77\%}}}
\put( 35,680){\makebox(0,0)[lb]{\raisebox{0pt}[0pt][0pt]{\footnotesize 54\%}}}
\put( 95,590){\makebox(0,0)[lb]{\raisebox{0pt}[0pt][0pt]{\footnotesize 79}}}
\put(175,590){\makebox(0,0)[lb]{\raisebox{0pt}[0pt][0pt]{\footnotesize 56}}}
\put(160,555){\makebox(0,0)[lb]{\raisebox{0pt}[0pt][0pt]{\footnotesize Generalization}}}
\end{picture}
\end{center}
\caption{Determiner Selection as Classification of Surface Sentences}
\label{surface}
\end{table}

\section{Applications in Multilingual NLP}

The main task of this paper has been to identify a set of semantic features
for the description of the definiteness category in English
and apply it to instances of plural nouns in a real text.
An application to grammar checking has been spelled out in the
former section. 
The results lead us to expect that with the development of a more
sophisticated textual coding, we may have a practical tool for
checking and correcting definiteness of English plural nouns.

The work reported here can also be used for multilingual interpretation and
generation. This is especially interesting
for languages without nominal determiners, such as Japanese or Russian.
In these cases other grammatical information that is provided in the surface
coding, e.g. Japanese particles with topic/comment contrast combining
the agentive/givenness dimensions and Ja\-pa\-ne\-se word order and nominal 
classifiers, can be used to set the semantic features of the 
intermediate, interlingual representation (cf. \cite{Wada94}).
Generation of an English determiner can then be handled by the 
unilingual learned generation function.
%

The history of machine translation and text understanding 
has shown that mere
surface scanning and textual matching approaches tend to level off
as they have no capacity for improving performance beyond that of
the statistical data analysis tool \cite{Nirenburg92}.
In contrast, using explicit semantic representations which can be linked to
cognitive models provides a basis for both human language understanding
and practical NLP.
Flat surface analysis may perform much better with huge data sets and less
information reduction. Still,
using semantic representations has additional advantages for interactive 
systems both for grammar checking and machine translation. The additional plane
of semantic representation allows a system to assess the validity of a given
decision and frame a question in other cases.

In order for the envisaged system to have real practical use
two kinds of additions are necessary (in addition to the general task
of improving the performance of the classifier):
\begin{itemize}
\item a textual encoding scheme that incorporates a method for coreference
resolution to set features in the dimension of anaphoric meaning reliably
\item a confidence measure for the proposed determiner, which would
make a remaining margin of error tolerable to a user.
\end{itemize}
The confidence measure could be composed of a value for the generation component which
would depend on the completeness of the semantic representation, and
a value for the analysis component, which would code the availability
of textual features and the probability values of the semantic feature
assignment (e.g. 0.9 or 0.6 ``collective'' etc.).

With these improvements the system could be a useful tool for anyone
who uses a foreign language and encounters frequent doubts of grammatical
correctness which no written grammar can answer:
* ``He answered me with the raised eyebrows'' is incorrect, but 
``with raised eyebrows'' or ``with the eyebrows raised in a mocking
twist'' is fine.
\appendix
\section{Appendix: Semantic dimensions and features}
\subsection{Generalized quantification}
\begin{itemize}
\item[1.] {\bf num}
quantifier with an explicit quantity, e.g. four, five etc.
\item[2.] {\bf unique}
a plural object may also be unique, for instance, 
{\it the arts}, {\it the London hospitals}
This is possible when it has a {\em collective} identity (s.below).
\item[3.] {\bf some}
an unspecified quantity, which  constitutes a small percentage
\item[4.] {\bf most}
an unspecified quantity, which constitutes a large percentage
\item[5.] {\bf all}
universal quantification, constrained with respect to the
discourse setting
\item[6.] {\bf general}
universal quantification, unconstrained with respect to discourse,
but pragmatically constrained 
\end{itemize}

\subsection{Anaphoric relation}
\begin{itemize}
\item[7.] {\bf given}
noun phrase with a co-referring antecedent
\item[8.] {\bf implied}
noun phrase which refers to an object implied by a lexical relation
\item[9.] {\bf new}
noun phrase that introduces a new referent
\end{itemize}

\subsection*{3.\hspace{5mm}Reference to Discourse Objects}
\begin{itemize}
\item[10.] {\bf denotation}
noun phrase that denotes an object term in discourse
(e.g., {\it He was walking about in the park})
\item[11.] {\bf predication}
noun phrase that denotes a property in discourse
(where a property is a one-place relation of a discourse object)
(e.g., {\it It's more a park than a garden})
\end{itemize}

\subsection*{4.\hspace{5mm}Boundedness}  
\begin{itemize}
\item[12.] {\bf mass}
reference to an unbounded quantity of one kind
(e.g., {\it a Lovely Young Thing with tight poodle CURLS})

\item[13.] {\bf pieces}
reference to a collection of individuals
(e.g., {\it 
Those dreadful policewomen in funny HATS who bother people in parks!} )
\end{itemize}

\subsection*{5.\hspace{5mm}Agentive involvement}
\begin{itemize}
\item[14.] {\bf collective} a plural noun referring to set of individuals and a
common action (e.g., {\it The two girls sang a duet.})
\item[15.] {\bf distributive}
a plural noun referring to a set of objects and individual actions
(e.g., {\it Four people brought a salad to the party.}) 
\end{itemize}


\begin{thebibliography}{}

\bibitem[\protect\citeauthoryear{Aone and Bennett}{1996}]{Aone-Bennett96}
Chinatsu Aone and Scott~William Bennett.
\newblock Applying machine learning to anaphora resolution.
\newblock In Stefan Wermter, Ellen Riloff, and Gabriele Scheler, editors, {\em
  Learning for natural language processing: Statistical, connectionist and
  symbolic approaches}, Lecture Notes in Artificial Intelligence. Springer,
  1996.

\bibitem[\protect\citeauthoryear{Bauer}{1995}]{BauerDIPLOM}
Stefan Bauer.
\newblock Entwicklung eines {E}ingabe-{T}aggers f{\"u}r
  lexikalisch-syntaktische {I}nformation.
\newblock Master's thesis, Technische Universit{\"a}t M{\"u}nchen, November
  1995.

\bibitem[\protect\citeauthoryear{Brill}{1993}]{Brill93}
Eric Brill.
\newblock {\em A Corpus-Based Approach to Language Learning}.
\newblock PhD thesis, University of Pennsylvania, Department of Computer and
  Information Science, 1993.

\bibitem[\protect\citeauthoryear{Church}{1988}]{Church88}
Kenneth~W. Church.
\newblock A stochastic parts program and noun phrase parser for unrestricted
  text.
\newblock In {\em Proceedings of the Second Conference on Applied Natural
  Language Processing}, pages 136--143, 1988.

\bibitem[\protect\citeauthoryear{Connolly \bgroup \em et al.\egroup
  }{1995}]{Connolly95}
Dennis Connolly, John~D. Burger, and David~S. Day.
\newblock A machine learning approach to anaphoric reference.
\newblock In D.~Jones, editor, {\em Learning for Natural Language Processing}.
  University Collecge London, 1995.

\bibitem[\protect\citeauthoryear{Fahlman}{1988}]{Fahlman88}
Scott Fahlman.
\newblock Faster-learning variations on back-propagation: An empirical study.
\newblock In T.~Sejnowski, G.~Hinton, and D.S. Touretzky, editors, {\em
  Proceedings of the 1988 Connectionist Models Summer School}. Morgan Kaufman,
  1988.

\bibitem[\protect\citeauthoryear{Jelinek}{1985}]{Jelinek85}
Fred Jelinek.
\newblock Markov source modeling of text generation.
\newblock In J.K. Skwirzinski, editor, {\em Impact of Processing Techniques on
  Communication}. Nijhoff, Dordrecht, 1985.

\bibitem[\protect\citeauthoryear{Kamp and Reyle}{1993}]{Kamp93}
Hans Kamp and Uwe Reyle.
\newblock {\em From Discourse to Logic: Introduction to Modeltheoretic
  Semantics of Natural Language, Formal Logic and Discourse Representation}.
\newblock Studies in Linguistics and Philosophy. Kluwer, 1993.

\bibitem[\protect\citeauthoryear{Link}{1991a}]{Link91a}
Godehard Link.
\newblock First order axioms for the logic of plurality.
\newblock In J.~Allgayer, editor, {\em Processing Plurals and Quantifications}.
  CSLI Notes, Stanford, 1991.

\bibitem[\protect\citeauthoryear{Link}{1991b}]{Link91b}
Godehard Link.
\newblock Plural.
\newblock In A.~von Stechow and D.~Wunderlich, editors, {\em Handbuch
  Semantik}. De Gruyter, 1991.

\bibitem[\protect\citeauthoryear{Miller and others}{1993}]{Miller93}
George~A. Miller et~al.
\newblock Introduction to {W}ord{N}et: An on-line lexical database.
\newblock Technical report, Princeton, 1993.

\bibitem[\protect\citeauthoryear{Nirenburg \bgroup \em et al.\egroup
  }{1992}]{Nirenburg92}
S.~Nirenburg, J.~Carbonell, M.~Tomita, and K.~Goodman.
\newblock {\em Machine Translation : A Knowledge-Based Approach}.
\newblock Morgan Kaufman, 1992.

\bibitem[\protect\citeauthoryear{Scheler and
  Schumann}{1995}]{Scheler/Schumann95}
Gabriele Scheler and Johann Schumann.
\newblock A hybrid model of semantic inference.
\newblock In Alex Monaghan, editor, {\em Proceedings of the 4th International
  Conference on Cognitive Science in Natural Language Processing (CSNLP 95)},
  pages 183--193, 1995.

\bibitem[\protect\citeauthoryear{Scheler}{1994}]{Scheler94g}
Gabriele Scheler.
\newblock Extracting semantic features for aspectual meanings from a syntactic
  representation using neural networks.
\newblock Technical Report FKI-191-94, Institut f{\"u}r Informatik, Technische
  Universit{\"a}t M{\"u}nchen, May 1994.

\bibitem[\protect\citeauthoryear{Scheler}{1996}]{Scheler96}
Gabriele Scheler.
\newblock Generating {E}nglish plural determiners from semantic
  representations.
\newblock In S.~Wermter, E.~Riloff, and G.~Scheler, editors, {\em Learning for
  natural language processing: Statistical, connectionist and symbolic
  approaches}, pages 61--74. Springer, 1996.

\bibitem[\protect\citeauthoryear{Schmid}{1994}]{Schmid94}
Helmut Schmid.
\newblock Part-of-speech tagging with neural networks.
\newblock In M.~Nagao, editor, {\em Proceedings of COLING}, pages 172--176,
  Kyoto, 1994.

\bibitem[\protect\citeauthoryear{Thompson and
  Martinet}{1969}]{Thompson/Martinet}
A.J. Thompson and A.V. Martinet.
\newblock {\em A Practical English Grammar}.
\newblock Oxford University Press, 1969.

\bibitem[\protect\citeauthoryear{Wada}{1994}]{Wada94}
H.~Wada.
\newblock A treatment of functional definite descriptions.
\newblock In M.~Nagao, editor, {\em Proceedings of COLING}, pages 789--795,
  Kyoto, 1994.

\bibitem[\protect\citeauthoryear{Yarowsky}{1992}]{Yarowsky92}
David Yarowsky.
\newblock Word sense disambiguation using statistical models of roget's
  categories trained on large corpora.
\newblock In {\em Proceedings of COLING 1992}, pages 454--460, 1992.

\bibitem[\protect\citeauthoryear{Zell and others}{1993}]{SNNS}
Andreas Zell et~al.
\newblock {\em Snns User Manual v. 3.1}.
\newblock Universit\"at Stuttgart: Institute for parallel and distributed
  high-performance systems, 1993.

\end{thebibliography}
\end{document}